# Knowledge Management in the Companion Cognitive Architecture

**Constantine Nakos**                                     CNAKOS@U.NORTHWESTERN.EDU
**Kenneth D. Forbus**                                     FORBUS@NORTHWESTERN.EDU
Qualitative Reasoning Group, Northwestern University, 2233 Tech Drive, Evanston, IL 60208 USA

## Abstract

One of the fundamental aspects of cognitive architectures is their ability to encode and manipulate knowledge. Without a consistent, well-designed, and scalable knowledge management scheme, an architecture will be unable to move past toy problems and tackle the broader problems of cognition. In this paper, we document some of the challenges we have faced in developing the knowledge stack for the Companion cognitive architecture and discuss the tools, representations, and practices we have developed to overcome them. We also lay out a series of potential next steps that will allow Companion agents to play a greater role in managing their own knowledge. It is our hope that these observations will prove useful to other cognitive architecture developers facing similar challenges.

## 1. Introduction

Knowledge is the fuel for cognitive architectures. Whether the task is question answering (Crouse, 2021), moral reasoning (Dehghani, 2009; Olson & Forbus, 2023), visual understanding (Chen, 2023), or game playing (Hancock, Forbus, & Hinrichs, 2020; Hancock & Forbus, 2021; Hinrichs & Forbus, 2011), cognitive architectures need a way to encode and manipulate knowledge about the world around them. The specific format will depend on the design of the architecture, its core claims, and the way it is deployed. Each architecture will also develop its own best practices for authoring, storing, and accessing the knowledge it uses, and these practices will evolve over time as the architecture scales up, changes, and develops new capabilities.

The Companion cognitive architecture (Forbus & Hinrichs, 2017) is undergoing just such a transition. Recent work on knowledge extraction (Ribeiro & Forbus, 2021) has laid the groundwork for automatically acquiring world knowledge that can be reused across domains, while the ongoing SocialBot project, an online expansion of the information kiosk described by Wilson et al. (2019), has required new knowledge management capabilities to support Companions in their role as social agents. For the Companion architecture to reach the next stage of its development, it will need a new set of tools and practices for knowledge management, enabling the agent itself to play a larger role in its own learning.

With new developments on the horizon, we take this opportunity to document the Companion knowledge stack as it currently stands, centered around the challenges we have faced and how we have addressed them. We discuss three major challenges: 1) *knowledge representation*, the design problem of encoding knowledge in a format amenable to reasoning, 2) *knowledge access*, the practical problem of efficiently making the knowledge available to a running Companion, and

3) *knowledge management*, the problem of scaling the knowledge base (KB) past a fixed set of hand-authored facts and allowing the Companion to acquire and save knowledge on its own.

While our discussion is grounded in Companions, we expect that similar challenges will arise in any cognitive architecture that attempts to reason and learn at scale. Thus, our goal in this paper is not just to checkpoint the Companion knowledge stack, but to share lessons learned in its development and foster a discussion on how to equip cognitive architectures for the complexities of managing their own knowledge and learning.

The remainder of the paper is structured as follows. We begin with an overview of the Companion cognitive architecture, its design principles, and two applications that have driven recent changes to its knowledge stack. Next, we discuss the stack itself, reviewing its established capabilities and presenting its new additions. We ground our discussion in three challenges— knowledge representation, access, and management—that the stack has grown to address. Section 6 outlines next steps for the knowledge stack and how we expect it to change as Companions evolve. We conclude with an overview of related work and some closing thoughts.

## 2. Companions

The main purpose of the Companion cognitive architecture (Forbus & Hinrichs, 2017) is to learn how to build *software social organisms*—agents that can learn, reason, and interact the way people do—as a step towards achieving human-level AI (Forbus, 2016). In support of this, Companions are equipped with a wide range of capabilities, from natural input modalities such as sketching and language to Hierarchical Task Network (HTN) planning to a robust analogy stack for performing mapping (Forbus et al., 2016), retrieval (Forbus, Gentner, & Law, 1995), and generalization (Kandaswamy & Forbus, 2012).

Every Companion consists of several agents operating in parallel, each one responsible for handling different tasks. For example, the Interaction Manager can parse a question asked by the user, then pass the query off to the Session Reasoner to determine the answer. Other agents handle capabilities like sketch understanding (Forbus et al., 2011) or strategic reasoning (Hancock & Forbus, 2021). This architecture provides coarse-grained parallelism and allows the Companion's capabilities to be expanded as needed by creating new types of agents.

Each agent in a Companion has its own instance of the FIRE reasoning engine (Forbus et al., 2010). FIRE gives the agent access to the analogy stack, logical inference, abduction, procedural attachments, and a working memory for storing reasoning results. But its most important duty is interfacing with the KB, allowing the agent to retrieve knowledge seamlessly as part of reasoning.

Companions depend heavily on conceptual knowledge. Rather than starting with an empty KB and trying to learn from scratch, Companions use NextKB (Section 3.3 ) as an approximation of the kind of conceptual knowledge a human has access to. While the coverage of NextKB is necessarily imperfect, it has been sufficient to support a variety of reasoning and learning tasks, including question answering (Crouse, 2021), legal reasoning (Blass, 2023), and learning by reading (Ribeiro & Forbus, 2021).

Two particular applications illustrate how Companions' knowledge management needs have grown in recent years: learning by reading and the SocialBot. We discuss each in turn before shifting our attention to the Companion knowledge stack and the three broad challenges that have shaped its development.

## 2.1 Learning by Reading

Ribeiro & Forbus (2021) show how a Companion can learn commonsense knowledge from text. They use analogical techniques to extract the knowledge and a BERT-based classifier to filter out erroneous facts. For example, the system can extract the fact `(properPhysicalPartTypes (MaleFn Deer) Antler)` from the sentence "Male deer have antlers." The system can learn to extract new relations from just a few examples. When tested on Simple English Wikipedia, it was able to extract 976 commonsense facts across 58 relation types from 2,679 articles with a precision of 71.4% (Ribeiro, 2023).

The long-term goal for this research is to enable a Companion to learn on its own and apply what it has learned to other reasoning tasks. However, this requires a level of autonomy that Companions do not yet possess. One of the main bottlenecks is the Companion's ability to manage the knowledge it learns: organizing it for reasoning with FIRE, integrating it into NextKB so other Companions can benefit from what one Companion learns, and tracking the provenance of learned knowledge so the quality of different sources can be monitored and errors can be corrected. These concerns have helped motivate the additions to the Companion knowledge stack discussed in Section 5. .

## 2.2 SocialBot

The SocialBot is a Companion-powered bot for the Microsoft Teams[1] platform, an expanded, online version of the multimodal information kiosk reported in Wilson et al. (2019). The SocialBot automatically scrapes the Northwestern University Computer Science Department website for information about courses, faculty, and events, then uses that information to answer user questions on Teams. Users can also share their food and drink preferences and their interest in CS topics so the bot can potentially provide personalized event recommendations. Ongoing research will expand the range of feedback users can provide, improving the SocialBot's recommendations and filling in gaps in its knowledge.

The SocialBot has stretched the Companion knowledge stack in two ways. First, its nightly scrapes of the CS Department website require tracking different versions of the same knowledge (e.g., updated event listings) to prevent stale data from appearing in its KB and thus its answers to user questions. This was one of the motivations behind the creation of the provenance cache, discussed in Sections 5.1 and 5.2 .

Second, the SocialBot must manage the knowledge it learns from users: organizing it, storing it persistently, and ensuring that it is available when a user asks a question. These requirements motivated the creation of the Archivist, discussed in Section 5.3 , a Companion agent dedicated to saving learned knowledge and serving it to other agents that need it.

## 3. Challenge #1: Knowledge Representation

---

[1] https://teams.microsoft.com

### 3.1 Knowledge Representation Basics

The foundation of any reasoning system is knowledge representation. The Companion cognitive architecture follows in the footsteps of Cyc (Lenat, 1995), a multi-decade effort to encode common-sense knowledge to support machine reasoning. Specifically, Companions use CycL, a language based on higher-order predicate calculus that represents assertions as Lisp-style *s*-expressions. Properly formed assertions consist of a *predicate* and its arguments, each of which may be an atomic *constant* or *non-atomic term* denoting an entity or a nested assertion. For example, the English statement "Joe likes apples." might be represented as the CycL assertion `(likesType Joe (FruitFn AppleTree))`, where `likesType` is the predicate, `Joe` is the constant for Joe, and `(FruitFn AppleTree)` is the non-atomic term that represents the fruit (the *function* `FruitFn`) of an apple tree (`AppleTree`).

In this example, `(FruitFn AppleTree)` denotes the *collection* of all apples, whereas `Joe` denotes an *individual* entity. Individuals and collections form the backbone of the ontology. The two are linked through the `isa` predicate: `(isa Nero-TheCat Cat)` states that the individual `Nero-TheCat` is a member of the collection `Cat`. Collections are related to each other through the `genls`[2] predicate: `(genls Cat Mammal)` states that all members of the collection `Cat` are members of the collection `Mammal`. `genls` is transitive, so if `(genls Mammal Animal)`, then all `Cat`s are `Animal`s. Higher-order collections support the categorization of collections themselves, such as `(isa Cat BiologicalSpecies)`.

Knowledge is stored in KRF[3] files. Each file consists of a series of assertions to load into a running KB, along with processing directives that affect how they are loaded. (The directives are `in-microtheory`, discussed in Section 3.2, and `with-provenance`, discussed in Section 5.2.) KRF files also support a handful of macros that do not conform to CycL syntax. These are useful for encoding groups of related facts in a compact way. For example, HTN methods, preconditions, and preference rules are bundled together with the `defPlan` macro. Macros are expanded into valid CycL assertions before being loaded into the KB.

### 3.2 Microtheories

Not all knowledge belongs together. Conflicting knowledge about the world can lead to logical contradictions. Reasoning problems in one domain may be derailed by knowledge intruding from another domain, such as the inclusion of fictional characters in a real-world planning task. More broadly, knowledge must be maintained by humans, and thus must be organized in a way that facilitates common ontologizing tasks, such as lookup, knowledge entry, and revision.

To that end, a Companion's knowledge is organized into *microtheories* (MTs; Guha, 1992; Lenat, 1998), bundles of assertions that are logically consistent and belong in the same reasoning context. All KB queries and reasoning operations are *contextualized* with a microtheory that determines which assertions are visible to that operation. For example, two physics models can be stored in different microtheories and queried independently, even if they are incompatible with each other.

---

[2] For "generalization".

[3] For "knowledge representation file".

Microtheories are connected using `genlMt` statements. `(genlMt MT1 MT2)` means that `MT1` inherits the contents of `MT2`. Any query made in `MT1` will see the facts stored in both `MT1` and `MT2`. The `genlMt` relation is transitive, allowing microtheories to be arranged in a hierarchy. If `MT2` inherits facts from `MT3`, `MT1` will inherit those facts as well. `genlMt` is also monotonic. If a fact is true in `MT2`, it must also be true in `MT1` and cannot be overridden or retracted. Monotonicity greatly simplifies the problem of determining which facts are true in which contexts, but care must be taken to keep high-level microtheories clear of any facts that do not belong in every inheriting microtheory.

The primary way to specify the microtheory of an assertion is the `in-microtheory` directive in a KRF file. When a file is being loaded into the KB, `(in-microtheory FactsAboutJoeMt)` indicates that all subsequent facts should be stored in the microtheory `FactsAboutJoeMt`, up until the next `in-microtheory` statement. This groups the contents of each KRF file into one or more microtheories. The `ist-Information` predicate can be used to specify the microtheory of an individual fact, as in `(ist-Information FactsAboutJoeMt (likesType Joe (FruitFn AppleTree)))`.

Microtheories are invaluable from the perspective of both reasoning and ontologizing. In addition to keeping contradictory facts separate, microtheories greatly reduce the amount of work required for deduction. The reasoner does not have to try every rule in the KB, only the ones that are visible from the current reasoning context. The compartmentalization of knowledge makes it easy to mix and match different types of knowledge. For example, a scratchpad microtheory might inherit rules from one MT and facts from another, letting the system reason about the combination of the two with no added effort. Finally, extending the ontology just requires finding the location in the `genlMt` hierarchy where the new knowledge will have the right effect.

### 3.3 NextKB

With the knowledge representation format settled, the next question becomes what knowledge to represent. Companions uses NextKB[4], a knowledge base derived from OpenCyc and extended with linguistic knowledge, qualitative representations, Companions rules and plans, and more. NextKB serves as a seed for a Companion's learning and reasoning, providing it with broad common-sense knowledge about the world. NextKB consists of over 700,000 facts spread across 1,300 microtheories and 1,000 KRF files. The ontology contains 83,000 collections, 26,000 predicates, and 5,000 functions, with over 3,000 rules to support reasoning.

So far, the knowledge in NextKB comes from three sources: the original OpenCyc ontology, manual extensions written by Companion users, and extensions generated semi-automatically from existing resources. All of these require some degree of hand curation. In Sections 5 and 6, we discuss the shift towards automatic extension of NextKB and how our representation and storage schemes are changing to accommodate it.

## 4. Challenge #2: Knowledge Access

### 4.1 PlanB

---

[4] http://www.qrg.northwestern.edu/nextkb/index.html

Companions access the knowledge in NextKB using PlanB[5], a knowledge base module written in Common Lisp. PlanB serves as an interface to an underlying database containing the contents of NextKB, as well as any additional KRF files the Companion has loaded or facts it has stored. The FIRE reasoning engine acts as a bridge between the Companion agent and PlanB so that, during inference, the KB can be queried seamlessly alongside other sources of knowledge. The details of PlanB's implementation are discussed in the following sections.

It is important to note that, while PlanB is part of a cognitive architecture, it does not embody any specific psychological claims. PlanB acts as a Companion's long-term memory, but it makes no attempt to model human memory except in the coarsest functional terms. Not every fact stored in the KB has a human analogue (i.e., some are only maintained for bookkeeping), and questions of representation are often settled on a practical basis, not a psychological one. We leave any psychological claims about memory to specific models implemented on top of PlanB, such as the MAC/FAC (Forbus et al., 1995) model of analogical retrieval.

## 4.2 AllegroCache

PlanB uses AllegroCache[6], a fast and scalable database that natively supports Lisp data types. In addition to typical database functionality such as persistent storage and rollback, AllegroCache offers two features of interest: a `persistent-class` Lisp metaclass that allows Lisp objects to be stored directly in the database, and the `map-range` class, which supports key-value storage.

`persistent-class` is the backbone of PlanB. Facts are stored as persistent objects with slots for their Lisp form, an integer ID, and the microtheories in which they are believed. KB concepts are stored as *conceptual entities*, persistent objects that can be specialized according to the type of the concept. For example, the subclass for predicates has extra slots for storing the predicate's arity, whether it is commutative, and so on. Like facts, conceptual entities can be looked up by their Lisp form (*s*-expression) or their integer ID.

`map-ranges` also play an important role in PlanB. They function as persistent hash tables whose keys can be any basic Lisp data type. Internally, the keys are sorted in a way that makes retrieval of sequential keys efficient using cursors. `map-ranges`' flexibility makes them useful for any data that does not fit neatly into the `persistent-class` paradigm, such as the mentions map or special facts (both discussed below).

Testing has revealed that `map-ranges` are more efficient than `persistent-classes`. As the usage patterns for PlanB stabilize over time, we are planning to shift more PlanB functionality to `map-ranges`, trading the convenience of `persistent-classes` for improved speed. The shift will have the added benefit of making PlanB more portable, as AllegroCache's `map-range` is closer to the kind of functionality supported by other database systems.

## 4.3 Retrieval

The primary function of a knowledge base is to retrieve stored facts on demand. PlanB supports unification-based retrieval. Given a query pattern with zero or more variables, PlanB will retrieve all facts which unify with the pattern. Queries can be contextualized with a microtheory, in which case only facts believed in that microtheory (either directly or by `genlMt` inheritance) will be returned.

---

[5] So named because it started as an alternative to an earlier system that turned out to be less performant.
[6] https://franz.com/products/allegrocache/

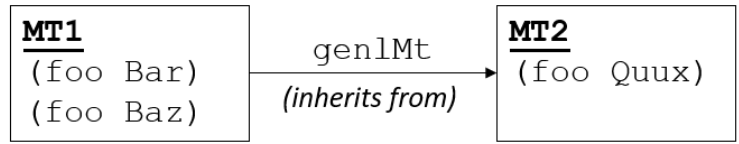

*Figure 1.* Sample KB with microtheories.

Consider the KB shown in Figure 1. Querying (`foo Bar`) in `MT1` will retrieve (`foo Bar`) but nothing else. The query is *ground* (i.e., it has no variables), so no other facts can unify with it. Querying (`foo ?x`) in `MT1`, where `?x` is a variable, will retrieve (`foo Bar`), (`foo Baz`), and (`foo Quux`), the latter inherited from `MT2`. Lastly, querying (`foo ?x`) in `MT2` will only retrieve (`foo Quux`) because the other facts are only visible from `MT1`.

PlanB uses *coarse coding* (Forbus et al., 2010) to index its facts. Every fact is indexed according to the entities it mentions in a `map-range` known as the *mentions map*. Mentions are stored as entity-location-fact triples, where the location is the position of the argument in which the entity appears. Arguments can have nested structure, but only the top-level position matters for the purpose of calculating mentions (hence "coarse").

For example, (`foo (TheList A B)`) will have mentions for `foo` in position 0 and for `TheList`, `A`, and `B` in position 1. (`foo (TheList B A)`) will also have the same mentions, even though `B` and `A` have swapped places within the same position, and thus the expressions would not unify.

Mentions are compressed into integer keys using the entity and fact IDs, so that mentions for the same entity are stored adjacently. This makes it efficient to look up all mentions of an entity, or all mentions of the entity in a particular position, using `map-range`'s cursor mechanism. This is useful for automatic case construction to support analogical reasoning and learning.

During retrieval, PlanB breaks down the query pattern to mentions the same way. Variables are ignored. For each entity mention in the query pattern, PlanB uses the mentions map to retrieve a *bucket* of facts that mention the same entity in the same location. Because a fact will only unify with a pattern if it has all of its mentions, PlanB takes the intersection of these buckets to find a set of candidate facts that could unify with the query. The candidates that unify are then filtered by microtheory to produce the final set of retrieved facts.

### 4.4 Special Facts & Indexed Facts

PlanB supports the definition of *special facts*, facts which are stored and retrieved using custom handlers. Special facts are used when efficiency is paramount, or when the nature of the fact lends itself to a specific type of storage. For example, genlMt statements are stored in a persistent TMS that allows efficient computation of microtheory inheritance (Forbus et al., 2010). Other special facts include fact probability and provenance (discussed below), bookkeeping for analogical retrieval, and templates for converting KB concepts to natural language.

Special facts are retrieved much faster than regular facts, as they can take advantage of custom lookup strategies, but the need for specialized code for each type of fact makes them harder to maintain. Since indexing every argument is not always necessary or desirable, special facts bypass the mentions map, meaning they can only be accessed using predefined query patterns.

As a compromise between the two extremes, PlanB supports *indexed facts*, regular facts that have one of their arguments designated as a key. When PlanB receives a query whose predicate is indexed, if the argument corresponding to that predicate's key is ground, it short-circuits the usual retrieval process and looks up the key directly in a special index. For example, the lexicon entries used by CNLU (Tomai & Forbus, 2009), Companions' semantic parser, are indexed by the token argument, making the common use case of mapping a token to its lexicon entries highly efficient. For queries where the key argument is not specified, indexed facts are retrieved the same way as regular facts, using the mentions map.

## 5. Challenge #3: Knowledge Management

### 5.1 The Epistemic Layer

The knowledge organization scheme we have discussed so far can be broken down into two layers: the *physical layer* and the *logical layer*. The physical layer consists of the KRF files where knowledge is stored before being loaded into the KB.[7] The basic unit of organization is the file, but directory structure also matters, both for the developers tasked with expanding a Companion's knowledge and for the index files that determine which KRF files should be incorporated into a KB build.

The primary way changes to the physical layer are propagated is through version control. Updated KRF files are checked in to a repository, where other developers can pull and reload them, or else wait for a fresh build of the KB that will contain all of the latest updates. This setup works well for the way NextKB has been developed so far: hand-curated knowledge maintained by a small set of developers and only one KB configuration. In Section 6, we discuss how these circumstances are changing.

In contrast, the logical layer refers to knowledge as it exists in the KB. The basic unit of organization is the microtheory, which bundles knowledge into units that are useful for reasoning. The logical layer is independent of the physical layer. One file can have multiple microtheories, and the contents of one microtheory can be spread out across multiple files.

This independence is very useful. Files can be merged or split without changing the logical structure of the KB, and the KB need only concern itself with facts and microtheories, not the files themselves. However, the independence of the logical and physical layers also leads to a problem. Once read into the KB, there is no way to tell what file a fact came from, so there is no easy way to correct a fact after it has been stored. Iterative correction of KRF files is a core part of Companion development, making this a pain point. Common workarounds include forgetting an entire microtheory so that its contents can be loaded fresh (cumbersome for large MTs split across multiple files) or saving a copy of the last-loaded version of a KRF file so its facts can be forgotten before the corrected version is reloaded (an easy step to forget).

---

[7] Knowledge stored dynamically in the KB during a Companion session can also be considered part of the physical layer, but, given that it must eventually be exported to a KRF file to persist, we ignore it for now.

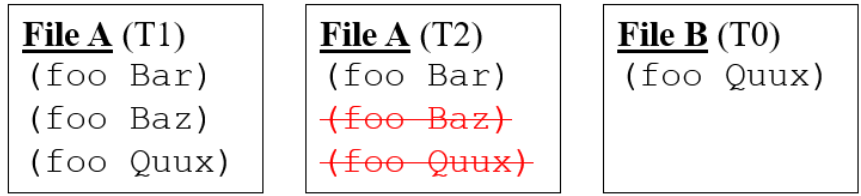

*Figure 2*. Two sample files, one with two versions. Deleted lines are in red.

This problem motivated the introduction of a new knowledge layer: the *epistemic layer*. Where the physical layer is organized by file and the logical layer is organized by microtheory, the epistemic layer is organized by *provenance*. Every fact that is loaded into the KB is tagged with metadata recording where it came from, namely the *provenance event* that caused the fact to be known. Typical provenance events include loading a KRF file, scraping a webpage, or having a conversation with a user. More generally, any meaningful occurrence that stores facts in the KB can be a provenance event.

Provenance events are represented in the KB as ordinary conceptual entities, so facts about them can be stored, retrieved, and reasoned with. Every provenance event has a *source*, where the information came from, and a *timestamp*, when it was provided. For example, when a KRF file is loaded, PlanB creates a provenance event that has the KRF file (represented as a URL) as its source and the file modification date (represented in Lisp universal time) as its timestamp. Subsequent loads of the same file would share the same source but have a different timestamp, depending on when the file was modified.

One upshot of this is that updating KRF files becomes effortless. The new version of the file is loaded into the KB, its facts are tagged with a new provenance event, and the old provenance event for the file is retracted. Any facts that were once supported by the old event but now have no support are forgotten.

Figure 2 shows a simple example. Suppose File A and File B have been loaded into the KB, and the facts from File A are tagged with provenance event `E1`, which has timestamp `T1`. Now suppose the user edits File A and reloads it, creating a new provenance event (`E2`, timestamp `T2`) and retracting the old one. `(foo Bar)` is in both versions of File A, so it remains in the KB, tagged with provenance event `E2`. `(foo Baz)` is not in the newer version of File A, so when `E1` is retracted, it will be forgotten. Finally, `(foo Quux)` has also been deleted from the new version of File A, but because it also appears in File B, it will not be forgotten.

Although the epistemic layer was initially introduced to address the problem of reloading KRF files, it has since proven useful in other ways. In the following section, we lay out further details of how PlanB handles provenance before moving on to applications of provenance as part of the broader Companion architecture.

## 5.2 The Provenance Cache

Provenance information is stored by PlanB in the *provenance cache*, a set of `map-ranges` that provide a lightweight mapping from facts to provenance events and vice versa. The `provenance` predicate interfaces with the provenance cache via special fact handlers, while

```
(with-provenance
    :source (MSTeamsUserFn "<user_id>")
    :timestamp (UniversalTimeFn 3919440252)
    :entity Discourse-3919440252-8397)
```

*Figure 3.* `with-provenance` statement for a conversation with a SocialBot user.

hooks in PlanB's storage and retrieval functions ensure the provenance cache is kept up to date as facts are added and removed from the KB.

While provenance facts can be stored in the KB individually, the typical way to tag provenance is a `with-provenance` directive in a KRF file. `with-provenance` statements allow one to automatically specify the provenance of all facts that follow them in the file, analogous to what `in-microtheory` does for context. For convenience, every KRF file is assumed to begin with an implicit `with-provenance` statement that has the file as its source and the file modification date as its timestamp. Thus, the default form of provenance, KRF files, is tracked automatically.

Figure 3 shows an example of a `with-provenance` statement for a conversation with a SocialBot user. The two main arguments are **:source** and **:timestamp**, which specify the source and timestamp of the provenance event, here the user ID (truncated for length) and the time of the conversation, respectively. The optional **:entity** argument allows the provenance event to be passed in explicitly, rather than using a new conceptual entity generated by PlanB. This is useful when more needs to be said about the event, in this case the user's conversation with the system. The optional **:type** argument specifies what collection the event should be added to, the default being `InformationTransferEvent`.

Two other arguments deserve special mention. The **:meta** argument controls whether the event is tagged as *meta-provenance*. Just as provenance is added to every fact stored, meta-provenance is added to every `provenance` fact stored, resulting in nested provenance statements that can capture multiple layers of reporting. For example, if a file contains a record of a conversation with a user, the provenance would be the conversation and the meta-provenance would be the file loading event. Meta-provenance allows us to reload the file as usual in that scenario. Without it, the facts would only be tagged with the conversation, not the file they came from. While PlanB allows arbitrary nesting of provenance statements, the **:meta** argument currently only supports two layers, which so far has been sufficient for our needs.

The **:update** argument has a more specific role. When true, the provenance event is treated as an update to its source. Any previous provenance event for that source will be retracted when a newer one is loaded. The main use of this feature is to allow multiple versions of the same knowledge to coexist in KRF files without conflicting in the KB.

For example, the SocialBot performs nightly scrapes of a computer science department events webpage so it can answer questions about those events. The scraped information is tagged with a `with-provenance` statement whose source is the week that block of information pertains to and whose timestamp is the time of the scrape. The **:update** argument means that, when the new files are loaded into the KB, only the most up-to-date information is retained without having to expunge the old files.

Provenance statements are contextualized, so the provenance cache can distinguish between the same fact being stored in different microtheories. Facts may also be supported by multiple provenance events, as seen in Figure 2. This allows the system to forget one KRF file while leaving the contents of an overlapping file intact. It also opens the door to a Companion accumulating redundant knowledge from multiple sources to gauge its reliability.

Finally, provenance can be set programmatically, so that a set of facts stored by the Companion are tagged with a particular provenance event. By default, every Companion session is one such event, so that any facts that the Companion stores during that session can be exported to a KRF file for archival purposes or retracted if necessary. This is a step towards having a Companion manage its own knowledge by examining what it has learned and deciding what to save and what to throw away.

### 5.3 The Archivist

There has been one other noteworthy development in the way Companions handle knowledge. The Archivist is a Companion agent tasked with tracking knowledge and serving it up to other agents during a session. It receives updates to KRF files from other agents, serves the updated files to agents upon request, and maintains a queue of changes to commit to a version control repository. The net result is that a Companion agent can gather new information during a session, propagate that information to other agents, and save that information persistently.

The Archivist was designed to keep knowledge up to date in the SocialBot. To handle conversations with multiple users in parallel, the SocialBot maintains a pool of Companion agents running on a cluster, each with its own FIRE reasoner and instance of the KB. At the end of a conversation, the agent sends any knowledge it learned to the Archivist for persistent storage. At the beginning of a new conversation, the agent queries the Archivist for any relevant updates that have been posted by other agents.

## 6. Next Steps

While developments like the provenance cache and the Archivist have expanded the knowledge management capabilities of Companions, we have only scratched the surface of what is possible. The ultimate goal is a self-directed agent capable of acquiring and organizing knowledge on its own. In the remainder of this section, we discuss potential next steps towards this goal.

### 6.1 Self-Directed Learning

While prior work has explored how a Companion can direct its own learning (e.g. Hinrichs & Forbus, 2019), this capability has not been tested at scale. Ideally, a Companion should have a set of broad learning goals that drive its behavior. When not otherwise occupied, it should seek to satisfy its goals by reading text, asking a user questions, ruminating on previously acquired knowledge, or running games or simulations.

Supporting this behavior will require extending the Companion knowledge stack. Acquired knowledge must be vetted and stored persistently so that other Companions can benefit from what one of them has learned. The Archivist, as employed by the SocialBot, points the way, but the problem gets more complicated as the Companion's learning abilities grow.

Eventually, we will need another layer of knowledge organization: the *distribution layer*, which would govern how KRF files are packaged into KBs for use by various applications. The current setup focuses on a single, general-purpose KB: NextKB. But the existence of SocialBot user models—separate from the core NextKB distribution, subject to their own privacy constraints, and applicable only to a single domain—opens the door to other specialized bodies of knowledge. Being able to tag groups of KRF files for inclusion in or exclusion from specific KB builds is a logical next step once a Companion can gather more knowledge on its own.

## 6.2 Knowledge Integration & Trust

One of the more interesting problems that arise as a Companion takes responsibility for more of its learning is figuring out how to integrate knowledge from multiple sources. Knowledge from an external source may not necessarily be accurate, and it may not be consistent with what the Companion has learned from other sources. To robustly learn from webpages, books, interactions with humans, and/or other outside sources of information, a Companion must be equipped with ways to compare, assess, and deploy the knowledge it has learned.

One important aspect of this problem is trust. Not all knowledge sources are trustworthy. Users may speak in ignorance or with intent to deceive, webpages vary wildly in terms of their veracity, and, in the extreme case, large language models (LLMs) generate plausible-sounding text with no specific grounding in reality. As such, a Companion needs to build up a model of how trustworthy its sources are so it can reconcile contradictions, proactively vet knowledge from suspect sources, and prioritize learning from more reliable ones.

The provenance cache provides a basis for this, tagging the source of each fact in the KB and making it available for introspection. When the Companion detects a suspect fact in the course of its reasoning, it can note the source that was to blame and update its trust model accordingly.

Dempster-Shafer Theory (Shafer, 1976) is one option for tracking reliability. The theory lays out how to combine claims from different sources, producing belief and plausibility estimates that provide lower and upper bounds on how probable the argument from evidence to conclusion is. Olson & Forbus (2021) show how Dempster-Shafer Theory can be used to resolve competing testimony about norms. Extending this to general learned knowledge is an avenue of future work.

## 6.3 Extending the Provenance Cache

While the current iteration of the provenance cache has proven useful, there are several additions that will help it support Companions' learning. First, the provenance cache should support events with hierarchical structure, such as scraping a single webpage as part of a larger scraping pass or running one trial within a larger experiment. The more information a Companion has about the structure of its experiences, the better equipped it will be to optimize its learning.

Second, the provenance ontology should be expanded to track different types of provenance events and sources. The current ontology consists of a handful of concepts in NextKB, which has been sufficient so far but lacks the systematicity of an ontology like PROV (Moreau et al., 2015).

Finally, tracking KB deletions in the provenance cache would help with replicability across Companion sessions. Right now, the cache only tracks when facts are stored, not when they are forgotten. Tracking deletions would allow one Companion to reproduce the KB changes made by another and would provide an automatic record when erroneous facts are forgotten.

## 7. Related Work

### 7.1 Knowledge in Cyc

Cyc shares the Companion cognitive architecture's focus on conceptual knowledge, and due to the massive size of its KB, its need for knowledge management is even greater. As such, Cycorp has developed a robust knowledge entry pipeline for the Cyc KB (A. Sharma, personal communication, March 10, 2024; Siegel et al., 2004). Ontological engineers work locally through a UI, and transcripts of their changes are sent to a central server for incorporation into the nightly KB build.

Notably, the Cyc build process is incremental, beginning with the most recent version of the KB and applying accumulated changes until the KB is up to date. In contrast, NextKB is rebuilt from scratch using standalone KRF files. Extending the provenance cache to track deletions will allow Companions to adopt Cyc-style updates as necessary by writing out the KB changes made during a session and using them to patch another KB instance.

The Cyc KB tracks the provenance of its assertions, including the ontologist who added them, the date and time they were added, and the original source of the information. Records of deletions are not present in the KB but can be derived from transcript files. This is in line with the aims of the Companion provenance cache, and as we plan our next steps, we take inspiration from Cyc's ability to show the user the original sources of facts used during inference.

### 7.2 Knowledge in Soar

The Soar cognitive architecture (Laird, 2012) provides an interesting contrast to the Companion knowledge stack. Where the Companion cognitive architecture prioritizes conceptual knowledge and reasoning, Soar prioritizes procedural knowledge and skill learning. This focus allows Soar to make concrete psychological claims about human performance and procedural learning while maintaining an architecture both general and performant enough for real-world applications.

The difference in emphasis leads to a variety of differences in knowledge representation and storage. Soar represents declarative knowledge with graph structures. Working memory consists of a single connected graph, while long-term declarative memory stores multiple graphs that can be retrieved into working memory based on cues. Soar's declarative memory is partitioned into semantic memory, which contains general knowledge about the world, and episodic memory, which records the agent's experiences, saved as snapshots of previous working memory states.[8]

In contrast, Companions treat CycL assertions as the basic unit of declarative knowledge, have a single long-term memory organized into microtheories, and lack a single, system-wide, automatic mechanism for episodic memory (cf. Forbus & Kuehne, 2007; Hancock, Forbus, &

---

[8] Soar also has procedural and perceptual memories, but they are not relevant to the current discussion.

Hinrichs, 2020; Hancock & Forbus, 2021). Episodes of interest must be stored deliberately on a domain-specific basis, although provenance tracking opens the door to more open-ended episodic memory by organizing facts that are saved during a Companion session.

Both Soar and Companions have mechanisms for retrieving long-term memory elements based on partial descriptions. In Soar, retrieval must be invoked deliberately by an operator as part of a problem-solving strategy, whereas Companions treat retrieval as an ubiquitous part of planning and reasoning. Companions also support analogical retrieval (Forbus et al., 1995), where cases are retrieved based on their structural similarity to a probe.

### 7.3 Provenance

There is a large body of existing work on representing the provenance of information, especially for the Semantic Web (see Herschel, Diestelkämper, & Ben Lahmar (2017), Pérez, Rubio, & Sáenz-Adán (2018), and Sikos & Philp (2020) for reviews). Languages such as PML (Pinheiro da Silva, McGuinness, & McCool, 2003) and PML 2 (McGuinness et al., 2007) can encode explanations for the conclusions produced by reasoning systems, including the provenance of the information used, while the PROV family of documents (Moreau et al., 2015) lays out a formal specification for encoding different types of provenance.

These endeavors are more ambitious than the Companion provenance cache, and they serve a different purpose. Unlike applications that use the Semantic Web, Companions perform most of their reasoning locally using knowledge aggregated at a central repository. Provenance matters for managing knowledge within the KB, tracking the sources of learned knowledge, and assessing their reliability, but it plays much less of a role than in the Semantic Web, where reasoning is frequently distributed and knowledge is drawn *ad hoc* from a variety of sources. Consequently we have opted for a lightweight provenance ontology rather than replicating the full complexity of, say, PROV. However, as Companions' provenance needs grow, we will expand the ontology as needed, drawing on existing work for inspiration. In particular, PROV's tripartite distinction between entities, agents, and activities might be useful for our needs.

## 8. Conclusion

One of the most important aspects of a cognitive architecture is how it deals with knowledge. In this paper, we have outlined the Companion knowledge stack, including the ways Companions represent, access, and manage their knowledge. We have also discussed potential next steps for the architecture that will enable Companions to play a greater role in their own learning. It is our hope that these observations will prove useful to cognitive architecture developers facing similar challenges.

## Acknowledgements

The authors would like to thank Abhishek Sharma for his helpful comments about Cyc, Madeline Usher for the design and implementation of the Archivist, Samuel Hill for the conversations that led to the creation of the provenance cache, and to the many QRG members who have contributed to NextKB, PlanB, and the entire Companion knowledge stack over the years. The authors would also like to thank the reviewers for their insightful comments. This research was sponsored by the US Office of Naval Research under grant #N00014-20-1-2447 and by the US Air Force Office of Scientific Research under award number FA95550-20-1-0091.

## References

Blass, J. (2023). *Extracting and applying legal rules from precedent cases*. Doctoral dissertation, Department of Computer Science, Northwestern University, Evanston, IL.

Chen, K. (2023). *Visual understanding using analogical learning over qualitative representations.* Doctoral dissertation, Department of Computer Science, Northwestern University, Evanston, IL.

Crouse, M. (2021). *Question-answering with structural analogy*. Doctoral dissertation, Department of Computer Science, Northwestern University, Evanston, IL.

Dehghani, M. (2009). *A cognitive model of recognition-based moral decision making*. Doctoral dissertation, Department of Electrical Engineering and Computer Science, Northwestern University, Evanston, IL.

Forbus, K., Gentner, D., and Law, K. (1995). MAC/FAC: A model of similarity-based retrieval. *Cognitive Science*, *19*, 141-205.

Forbus, K., and Kuehne, S. (2007). Episodic memory: A final frontier (Abbreviated version). *Proceedings of AI for Interactive Digital Entertainment (AIIDE07)*. Palo Alto, CA.

Forbus, K., Hinrichs, T., de Kleer, J., and Usher, J. (2010). FIRE: Infrastructure for experience-based systems with common sense. *AAAI Fall Symposium on Commonsense Knowledge*. Arlington, VA.

Forbus, K., Usher, J., Lovett, A., Lockwood, K., and Wetzel, K. (2011). CogSketch: Sketch understanding for cognitive science research and for education. *Topics in Cognitive Science*, *3*, 648-666.

Forbus, K. (2016). Software social organisms: Implications for measuring AI progress. *AI Magazine*, *37*, 85-90.

Forbus, K. D., Ferguson, R. W., Lovett, A., and Gentner, D. (2016). Extending SME to handle large-scale cognitive modeling. *Cognitive Science*, *41*, 1152-1201.

Forbus, K.D. & Hinrichs, T. (2017) Analogy and qualitative representations in the Companion cognitive architecture. *AI Magazine*, *38*, 34-42.

Guha, R. V. (1992). *Contexts: A formalization and some applications*. Doctoral dissertation, Department of Computer Science, Stanford University, Stanford, CA.

Hancock, W., Forbus, K., & Hinrichs, T. (2020). Towards qualitative spatiotemporal representations for episodic memory. *Proceedings of the 33rd International Workshop on Qualitative Reasoning*. Santiago de Compostela, Spain.

Hancock, W. & Forbus, K. (2021). Qualitative spatiotemporal representations of episodic memory for strategic reasoning. *Proceedings of the 34th International Workshop on Qualitative Reasoning*. Montreal, Canada.

Herschel, M., Diestelkämper, R., & Ben Lahmar, H. (2017). A survey on provenance: What for? What form? What from?. *The VLDB Journal*, *26*, 881-906.

Hinrichs, T. and Forbus, K. (2011). Transfer learning through analogy in games. *AI Magazine*, 32(1), 72-83.

Hinrichs, T. and Forbus, K. (2019). Experimentation in a model-based game. *Proceedings of the Seventh Annual Conference on Advances in Cognitive Systems.* Cambridge, MA.

Kandaswamy, S. and Forbus, K. (2012). Modeling learning of relational abstractions via structural alignment. Proceedings of the 34th Annual Conference of the Cognitive Science Society (CogSci). Sapporo, Japan.

Laird, J. E. (2012). *The Soar cognitive architecture*. Cambridge, MA: MIT Press.

Lenat, D. B. (1995). CYC: A large-scale investment in knowledge infrastructure. *Communications of the ACM*, *38*, 33-38.

Lenat, D. (1998). *The dimensions of context-space* (Technical report). Cycorp, Inc., Austin, TX.

McGuinness, D. L., Ding, L., Da Silva, P. P., & Chang, C. (2007). PML 2: A modular explanation interlingua. *ExaCt 2007* , 49-55.

Moreau, L., Groth, P., Cheney, J., Lebo, T., & Miles, S. (2015). The rationale of PROV. *Journal of Web Semantics*, *35*, 235-257.

Olson, T. & Forbus, K. (2021). Learning norms via natural language teachings. *Proceedings of the 9th Annual Conference on Advances in Cognitive Systems 2021.* Virtual.

Olson, T. & Forbus, K. (2023). Mitigating adversarial norm training with moral axioms. *Proceedings of the Thirty-Seventh AAAI Conference on Artificial Intelligence*. Washington, DC.

Pinheiro da Silva, P., McGuinness, D. L., & McCool, R. (2003). Knowledge provenance infrastructure. *IEEE Data Engineering Bulletin*, *26*, 26-32.

Pérez, B., Rubio, J., & Sáenz-Adán, C. (2018). A systematic review of provenance systems. *Knowledge and Information Systems*, *57*, 495-543.

Ribeiro, D. & Forbus, K. (2021). Combining analogy with language models for knowledge extraction. *Proceedings of the Third Conference on Automatic Knowledge Base Construction.* Virtual.

Ribeiro, D. (2023). *Reasoning and structured explanations in natural language via analogical and neural learning.* Doctoral dissertation, Department of Computer Science, Northwestern University, Evanston, IL.

Shafer, G. (1976). *A mathematical theory of evidence*. Princeton, NJ: Princeton University Press.

Siegel, N., Goolsbey, K., Kahlert, R., & Matthews, G. (2004). *The Cyc system: Notes on architecture* (Technical report). Cycorp, Inc., Austin, TX.

Sikos, L. F., & Philp, D. (2020). Provenance-aware knowledge representation: A survey of data models and contextualized knowledge graphs. *Data Science and Engineering*, *5*, 293-316.

Tomai, E. and Forbus, K. (2009). EA NLU: Practical language understanding for cognitive modeling. *Proceedings of the 22nd International Florida Artificial Intelligence Research Society Conference*. Sanibel Island, FL.

Wilson, J., Chen, K., Crouse, M., C. Nakos, C., Ribeiro, D., Rabkina, I., Forbus, K. D. (2019). Analogical question answering in a multimodal information kiosk. *Proceedings of the Seventh Annual Conference on Advances in Cognitive Systems.* Cambridge, MA.